\documentclass{article} % For LaTeX2e

\usepackage{iclr2026_conference,times}

\usepackage{amsmath,amsfonts,bm}

\def\eqref#1{equation~\ref{#1}}
\def\1{\bm{1}}

\DeclareMathAlphabet{\mathsfit}{\encodingdefault}{\sfdefault}{m}{sl}
\SetMathAlphabet{\mathsfit}{bold}{\encodingdefault}{\sfdefault}{bx}{n}

\usepackage{hyperref}
\usepackage{url}

\usepackage{verbatim}
\usepackage{times}
\usepackage{latexsym}
\usepackage{arydshln}
\usepackage{amsmath}
\usepackage[T1]{fontenc}
\usepackage{xcolor}
\definecolor{deepgreen}{RGB}{0,100,0}
\usepackage[utf8]{inputenc}
\usepackage{multirow} 
\usepackage{microtype}
\usepackage{booktabs}
\usepackage{inconsolata}
\usepackage{algorithm}
\usepackage{algpseudocode}
\usepackage{graphicx}
\usepackage{float}
\usepackage{subfig}
\usepackage{lipsum}
\usepackage{graphicx}
\usepackage{tcolorbox}
\usepackage{amsfonts}
\usepackage{amsmath}
\usepackage{wrapfig}
\tcbuselibrary{listings,breakable}
\newcommand{\yz}[1]{\textcolor{red}{[YZ: #1]}}

\title{WebSeer: Training Deeper Search Agents through Reinforcement Learning with Self-Reflection}

\author{Guanzhong He\quad  Zhen Yang\quad  Jinxin Liu\quad  Bin Xu\thanks{Corresponding author: xubin@tsinghua.edu.cn}\quad  Lei Hou\quad  Juanzi Li \\
        Tsinghua University \\}

\iclrfinalcopy % Uncomment for camera-ready version, but NOT for submission.
\begin{document}

\maketitle
\begin{abstract}
% Retrieval-Augmented Generation (RAG) enhances Large Language Models by grounding them in external knowledge. 

%\yz{search agents ....}

%\yz{tradtional search agents workflow }

%While reinforcement learning has been used to train agentic models capable of interactive retrieval, existing approaches remain limited in tool-use depth \yz{agents lack reflection mechanism} and fail to address the complexities of real-world web environments. 

Search agents have achieved significant advancements in enabling intelligent information retrieval and decision-making within interactive environments.
Although reinforcement learning has been employed to train agentic models capable of more dynamic interactive retrieval, existing methods are limited by shallow tool-use depth and the accumulation of errors over multiple iterative interactions.
In this paper, we present WebSeer, a more intelligent search agent trained via reinforcement learning enhanced with a self-reflection mechanism. Specifically, we construct a large dataset annotated with reflection patterns and design a two-stage training framework that unifies cold start and reinforcement learning within the self-reflection paradigm for real-world web-based environments, which enables the model to generate longer and more reflective tool-use trajectories. 
Our approach substantially extends tool-use chains and improves answer accuracy. Using a single 14B model, we achieve state-of-the-art results on HotpotQA and SimpleQA, with accuracies of 72.3\% and 90.0\%, respectively, and demonstrate strong generalization to out-of-distribution datasets. %Our results highlight the importance of explicit training for tool-use depth and reflective reasoning in real-world retrieval scenarios. 
The code is available at \href{https://github.com/99hgz/WebSeer}{https://github.com/99hgz/WebSeer}.

\end{abstract}

\section{Introduction}

Large language models (LLMs) have demonstrated remarkable performance across a wide range of natural language processing tasks, including question answering, summarization, and dialogue generation~\citep{hendryckstest2021,rein2024gpqa}.
\iffalse
\yz{LM limits must use RAG and agent search}

\yz{compared with RAG, agent search benifits --  agentic ....}

\yz{fig1 -- demo case }

\yz{agent search limitations}

However, relying solely on the parametric knowledge of language models poses fundamental limitations: it is static, often outdated, prone to hallucinations~\citep{sardana2025real}, and it struggles to adapt to dynamic or domain-specific contexts.
To overcome these issues, \citep{lewis2021retrievalaugmentedgenerationknowledgeintensivenlp} proposed the Retrieval-Augmented Generation (RAG) framework, enabling models to retrieve external documents.
While effective, most RAG architectures follow rigid linear or tree-structured workflows~\citep{asai2023selfraglearningretrievegenerate,trivedi2023interleavingretrievalchainofthoughtreasoning,yu2024rankragunifyingcontextranking}, limiting the model’s ability to dynamically control retrieval, reflect on intermediate results, or revise earlier decisions—thus underutilizing the full reasoning capacity of LLMs.

Recent efforts have begun to treat retrieval as an external tool, allowing models to proactively invoke retrieval when needed~\citep{trivedi2023interleavingretrievalchainofthoughtreasoning,li2025searcho1agenticsearchenhancedlarge,jin2025searchr1trainingllmsreason}. This design empowers the model to follow more complex reasoning trajectories, mirroring how humans iteratively seek out and incorporate external information when solving problems. 
\fi
However, relying solely on the parametric knowledge of language models poses fundamental limitations: it is static, often outdated, and prone to hallucinations~\citep{sardana2025real}. To overcome these challenges, retrieval-augmented generation (RAG)~\citep{lewis2021retrievalaugmentedgenerationknowledgeintensivenlp}  approaches have been developed to enable models to access and retrieve external documents dynamically.

With the continuous advancement of model capabilities, agentic RAG~\citep{trivedi2023interleavingretrievalchainofthoughtreasoning,li2025searcho1agenticsearchenhancedlarge,jin2025searchr1trainingllmsreason} has emerged as a powerful paradigm. This design empowers models to follow more complex reasoning trajectories. Unlike traditional RAG systems~\citep{asai2023selfraglearningretrievegenerate,trivedi2023interleavingretrievalchainofthoughtreasoning,yu2024rankragunifyingcontextranking}, agentic RAG can freely browse vast knowledge sources available on the internet and leverage tools such as code execution to extend their skills, enabling them to tackle a more diverse range of tasks. 
While agentic RAG greatly extends the scope of tool use and demonstrates strong potential, existing approaches exhibits several notable limitations in practice. In particular, when faced with complex or open-domain tasks, current systems often struggle to maintain coherent reasoning chains and robust retrieval. Errors introduced at intermediate steps can easily accumulate, while the lack of effective coordination across components makes it difficult to achieve reliable end-to-end performance. Thus, despite its promise, agentic RAG still confronts a set of fundamental challenges that must be addressed.

% As shown in Figure~\ref{fig:fig0}, agentic RAG expands the scope of tool use, yet competing search agents struggle with shallow reasoning and error buildup, underscoring the challenges that remain.

% However, these approaches suffer from the following problems:
%\yz{1. ok 2. reflection 3. ok }

\begin{figure*}
    \centering
    \includegraphics[width=1\linewidth]{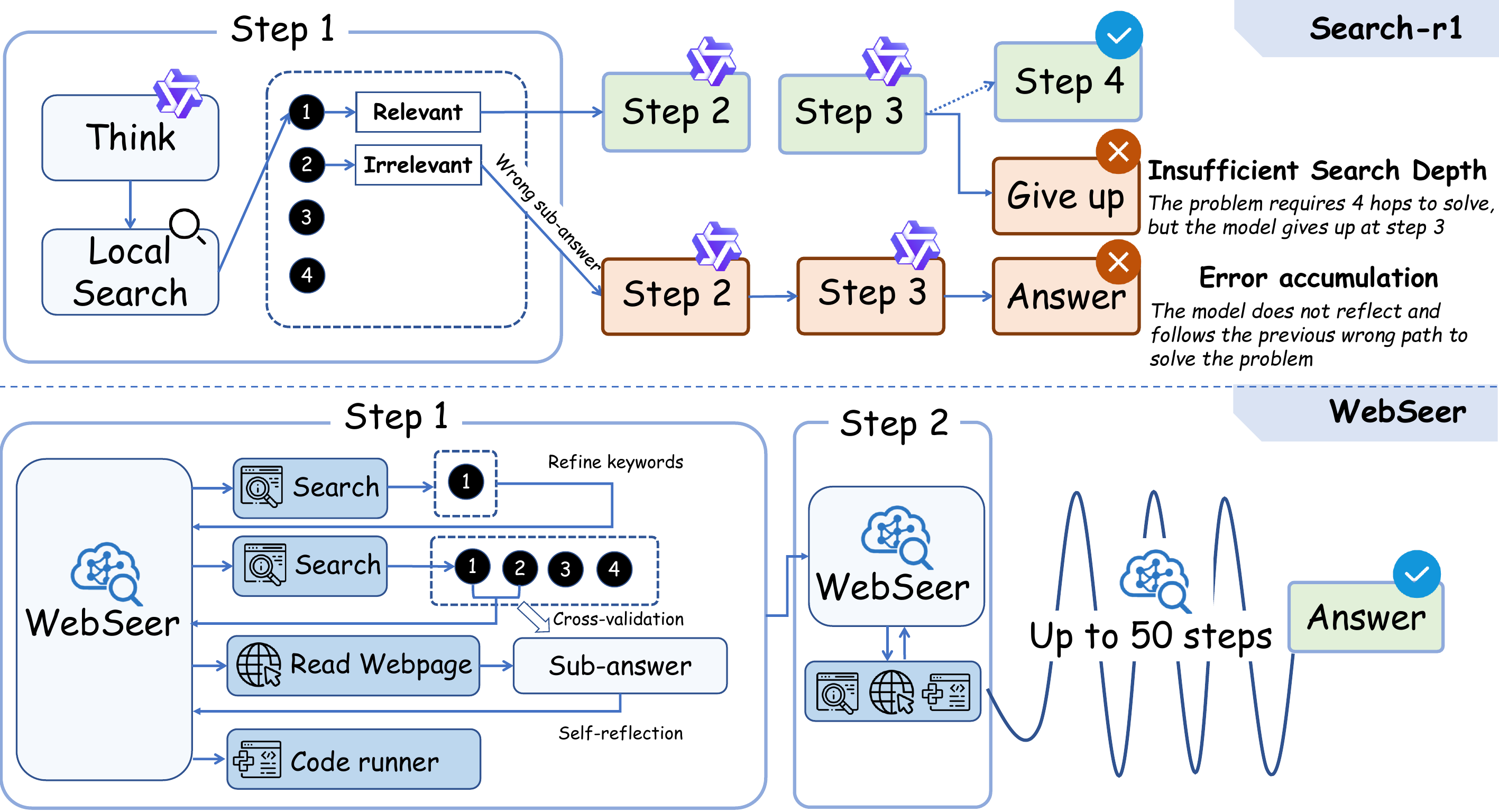}
    \caption{Comparasion between different search agent: Search-r1\citep{jin2025searchr1trainingllmsreason} faces shallow search and error buildup, while WebSeer trained using our self-reflection paradigm significantly improve tool calls and maintain rigorous reasoning.}
    \vspace{-2mm}
    \label{fig:fig0}
    \vspace{-5mm}
    % \vspace{-38pt} % 调整与下文距离
\end{figure*}

\textbf{1) Insufficient Search Calls.}
The most common issue when it comes to model invocation tools to solve problems is Insufficient API Calls~\citep{kokane2025toolscanbenchmarkcharacterizingerrors}, which is also often seen in the RAG scenario, as shown in the Appendix~\ref{app:cs1}. This may be due to models being biased toward synthesizing the currently available information into a plausible answer rather than actively seeking new or complementary knowledge. Existing work typically exhibits short tool-use chains~\citep{jin2025searchr1trainingllmsreason,song2025r1searcherincentivizingsearchcapability}, suggesting that models may fall into suboptimal behaviors—prematurely producing an answer instead of verifying its correctness or exploring further evidence.

\textbf{2) Lack of Spontaneous Self-Reflection Mechanisms.}
Current search agents lack spontaneous reflection steps in RAG scenario: models neither actively cross-verify information nor autonomously rewrite queries or backtrack retrievals when uncertain. As a result, if initial retrieval is incomplete, the generation phase expands answers based on flawed or partial context, amplifying early errors. %Especially in multi-hop reasoning tasks, such biases propagate as subsequent steps treat erroneous information as “known facts,” leading to cumulative errors.

\textbf{3) Neglect of Real-World Web Scenarios.}
Most existing work focuses on retrieval from local vector databases, with limited attention to more complex and open-ended web agent scenarios.

%\yz{WebSeer -- what is ? to solve/do } 

%In this work, we present the first two-stage agent training paradigm designed for real-world web-based search environments. 
In this paper, we introduce Webseer, a novel search agent designed to tackle complex real-world multi-hop question answering tasks. 
% problems through deeper exploration and self-reflection mechanism.
Different from prior approaches~\citep{jin2025searchr1trainingllmsreason, zheng2025deepresearcherscalingdeepresearch}, WebSeer explicitly encourages deeper exploration and integrates a build-in self-reflection mechanism, enabling the model to backtrack, reformulate queries, and iteratively refine its reasoning process.
As shown in Figure~\ref{fig:fig0}, previous agentic RAG primarily rely on short tool-use chains and often terminate once a superficially plausible answer is formed. In contrast, WebSeer actively prolongs the search trajectory and incorporates reflection steps, and revises its queries when uncertainty is detected. This design enables Webseer to gather more comprehensive evidence, mitigate the accumulation of errors, and improve robustness in open-domian multi-hop reasoning. Specifically, we design a two-stage training framework that unifies cold start and reinforcement learning within the self-reflection paradigm. A central component of this framework is Self-Reflective Reinforcement Learning (SRRL), which leverages answer correctness signals during multi-turn interactions to more effectively encourage reflective behavior. To ground the model in realistic web scenarios, we equip it with three complementary tools: a web search API for external knowledge acquisition, a webpage reader for lightweight comprehension of web content, and a code executor for precise computation.

\iffalse
\yz{two-stage paradigm} 

\yz{integrate three tools}

\yz{unify cold start + RL in self-reflection paradigm}

\yz{specially, dataset }

To enable interaction with realistic web scenarios, we equip the model with a set of external tools, including a web search API, webpage browser, and code execution module.

To scale up the length and depth of tool-use chains in complex problem solving, we incorporate a reflection mechanism inspired by reasoning strategies in mathematical problem solving and further optimize this behavior using the GRPO training method.

%Specifically, after generating an initial answer, the model is prompted to verify its correctness and consider whether alternative answers have been overlooked.
\fi

To support this framework, we construct a high-quality dataset of long-horizon reasoning trajectories through rejection sampling. These trajectories contain multiple rounds of answer refinement and substantially longer tool-use chains compared to conventional dialogue datasets. This training framework significantly increases the average length of tool invocation chains while maintaining rigorous reasoning quality. Compared to prior work~\citep{zheng2025deepresearcherscalingdeepresearch}, all decisions and tool interactions are handled by a single model, eliminating the need for auxiliary agent controllers or stronger backbone models. %Moreover, our approach supports standard OpenAI-compatible tool interfaces.

Overall, our contributions are three-fold:
\begin{itemize}
    \item We propose the first unified two-stage training framework with a self-reflection paradigm, enabling a more intelligent search agent that improves search depth, breadth, and accuracy.
    \item We propose a novel SFT data synthesis method that encourages proactive tool invocation, resulting in significantly longer and more complex tool-use chains;
    \item Through extensive experiments, we demonstrate notable improvements in answer accuracy, achieving SOTA on HotpotQA with a 72.3\% accuracy and on simpleQA with a 90.0\% accuracy. Our proposed training framework shows explicit self-verification behaviors and strong generalization to OOD datasets.
\end{itemize}

\section{WebSeer}
\begin{figure*}
    \centering
    \includegraphics[width=1\linewidth]{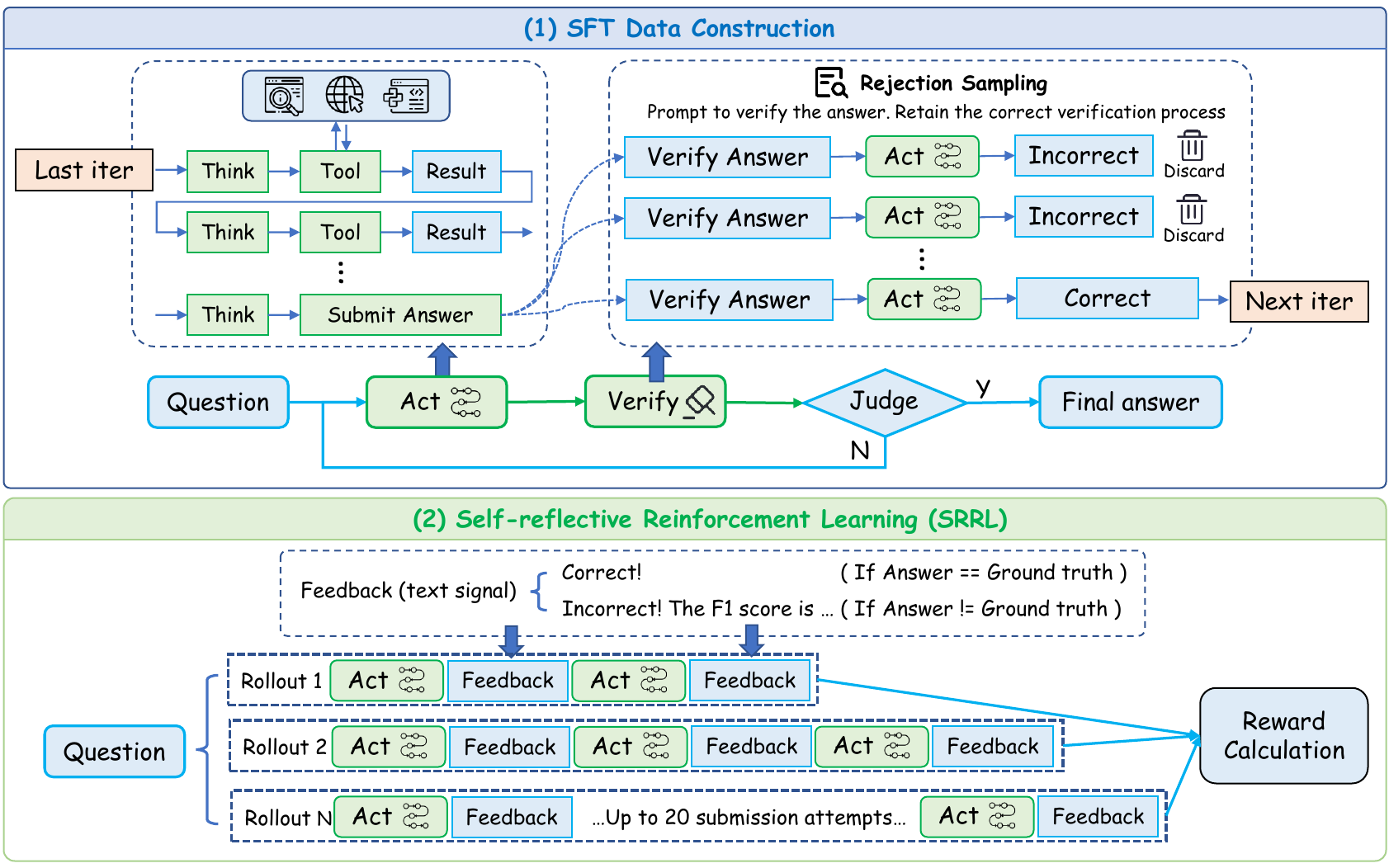}
    % \vspace{-4mm}
    \caption{Overview of our two-stage training framework. In the first stage, we construct self-reflective, tool-augmented reasoning trajectories. In the second stage, we apply SRRL, allowing the model to iteratively refine and resubmit answers with F\textsubscript{1}-based feedback.}
    \vspace{-3mm}
    \label{fig:pipeline}
    % \vspace{-18pt} % 调整与下文距离
    % \vspace{-2mm}
\end{figure*}

% \yz{22 Done!!!!}
We introduce WebSeer, a search agent for multi-hop question answering in real-world web settings. The model is trained with our proposed unified two-stage framework based on self-reflection as shown in Fig~\ref{fig:pipeline}. To obtain diverse high-quality data, we propose multi-turn rejection sampling for collecting positive trajectories used in supervised fine-tuning, enabling the model to learn reflective reasoning patterns. Building on this, we develop Self-Reflective Reinforcement Learning (SRRL), which incorporates feedback and preserves reflective context during RL, guided by effective reward design to achieve robust and optimal performance.

\subsection{Task Formulation}
For a given problem, our objective is to construct a reasoning chain augmented by tool invocations. Specifically, the reasoning chain consists of $N$ sequential steps, each comprising a model-generated output, a tool invocation, and the resulting observation. At each step, the model initially generates reasoning outputs based on the current context, then decides to invoke one or more external tools. The tool invocation parameters are extracted by the decoder, and subsequently, the invoked tools are executed, returning observations integrated into the reasoning context. The reasoning chain terminates if the model either abstains from further tool invocations or invokes a special submit-answer tool, signaling completion. Additionally, a predefined maximum step limit $T_{\text{max}}$ is enforced to prevent indefinite chaining, thus constraining the length of the reasoning trajectory.

\subsection{Tool Design}

In this work, we design three specialized tools to support external knowledge acquisition and reasoning execution. Additionally, we treat the Answer Submission step itself as a tool, ensuring that the model explicitly decides when to terminate reasoning and output the final answer.

\textbf{Search Engine}  
This tool receives a set of keywords and performs a Google search. It returns the top page of search results in a structured format, including the \textit{title}, \textit{URL}, and \textit{snippet} for each result. 

\textbf{Webpage Reader}  
Due to the prohibitive input length of raw HTML, we design the webpage reader to act as a lightweight, model-mediated summarization interface. Given a specific URL and a question, the system fetches the HTML content of the page and prompts the same language model to answer the question based on the page content. The tool then returns the model-generated answer.

\textbf{Code Executor}  
This tool accepts a Python code snippet, executes the code in a controlled environment, and returns the standard output as the tool result. 

\subsection{Self-reflective Reasoning Chain Construction}

Deepseek R1~\citep{deepseekai2025deepseekr1incentivizingreasoningcapability} has highlighted the importance of cold-start in reinforcement learning training. \citet{lee2025rearagknowledgeguidedreasoningenhances} provided several trajectories for training models to solve multi-hop QA problems. However, these trajectories only included cases where the correct answer was successfully found, without teaching the model how to handle situations in which an incorrect answer might arise. To address this limitation, we propose a multi-turn rejection sampling method to collect reasoning paths that incorporate reflective patterns. To overcome this limitation, we propose a multi-turn rejection sampling method that collects reasoning paths enriched with reflective patterns. To encourage trajectory diversity, we impose no restrictions on the form of reflection itself, retaining only those reasoning paths that ultimately converge to the correct solution.

Let $\mathcal{D}={(x_i, y_i^\ast)}_{i=1}^N$ denote a multi-hop QA dataset with ground-truth answers $y_i^\ast$. We consider two models: (i) a \emph{reasoner} $G$, which, given an instance and its interaction history, generates a tool-augmented reasoning path until producing a final answer; and (ii) an independent \emph{verifier} $V$, which assesses the factual correctness of a proposed answer by invoking the same tools and ultimately returning a judgment.
At reflection step $t \in \{1,\dots,n_{\max}\}$ for instance $x_i$, the reasoner receives the concatenated history 
$
H_{t-1} = \{P_1, R_1, \dots, P_{t-1}, R_{t-1}\},
$
and generates a tool-augmented path $P_t$ that culminates in an answer proposal
$
\hat{y}_i^{(t)} \;\leftarrow\; G(x_i, H_{t-1}).
$
To evaluate this proposal, we query the verifier up to a budget of $K$, yielding a verification outcome $R_t \sim V(\,\cdot \mid x_i, P_t, \hat{y}_i^{(t)}\,)$.

Each $R_t$ consists of (i) a binary judgment $J_t \in \{\textsc{correct}, \textsc{incorrect}\}$ regarding $\hat{y}_i^{(t)}$, and (ii) a tool-augmented path that may be appended to the full path.

\textbf{Validity predicate.}  
We define a predicate
$
\Psi\!\left(R_t, \hat{y}_i^{(t)}, y_i^\ast\right) \in \{0,1\}
$
that evaluates whether the judgment in $R_t$ \emph{matches factual correctness}, i.e.,
\[
\Psi\!\left(R_t, \hat{y}_i^{(t)}, y_i^\ast\right) =
\begin{cases}
1, & \text{if } \bigl(J_t=\textsc{correct} \ \wedge\  \hat{y}_i^{(t)}=y_i^\ast\bigr)\ \text{ or } \bigl(J_t^=\textsc{incorrect} \ \wedge\  \hat{y}_i^{(t)}\neq y_i^\ast\bigr),\\[4pt]
0, & \text{otherwise.}
\end{cases}
\]

So, if there exists
$\Psi=1$, we \emph{accept} the verifier’s feedback and update the history via concatenation,  
$
H_t \;\gets\; H_{t-1} \cup \{P_t, R_t\}.
$
Otherwise, we re-query the verifier to sample the next outcome
\[
R_t^{(m)} \sim V(\,\cdot \mid x_i, P_t, \hat{y}_i^{(t)}\,), \quad m=1,2,\dots,K,
\]
and evaluate $\Psi\!\left(R_t^{(m)}, \hat{y}_i^{(t)}, y_i^\ast\right)$.

If all $K$ attempts fail, then the instance $x_i$ is discarded, and we directly proceed to the next problem instance $x_{i+1}$.

\textbf{Iteration and termination.}  
The above process iterates for $t=1,2,\dots$ with the updated history until one of the following conditions holds:
\[
\textbf{(Success)} \quad \hat{y}_i^{(t)} = y_i^\ast \ \text{and} \ J_t=\textsc{correct},
\]
in which case we halt and record the finalized, verified trajectory
$
\mathcal{T}_i \;=\; \{P_1, R_1, P_2, R_2, \dots, P_t, R_t\},
$
\[
\textbf{(Budget stop)} \quad t = n_{\max},
\]
in which case the instance is not recorded as a successful trajectory. 

\textbf{Supervised fine-tuning.}  
Let $\{\mathcal{T}_i\}_{i=1}^N$ denote the all set of successful trajectories, where each trajectory $\mathcal{T}_i$ is represented as a token sequence
$
\mathcal{T}_i = \{ y^{(i)}_1, y^{(i)}_2, \dots, y^{(i)}_{T_i} \}.
$
To stabilize subsequent reinforcement learning and mitigate degenerate exploration, we perform supervised fine-tuning (SFT) of the model parameters $\theta$ on the dataset $\{(x_i, \mathcal{T}_i)\}_{i=1}^N$. Following empirical findings on iterative search training~\citep{zhang2025evolvesearchiterativeselfevolvingsearch}, we adopt a masked autoregressive negative log-likelihood (NLL) objective that excludes external observation tokens from the loss.

Let $\mathcal{O} \subset \mathcal{T}$ denote the subsequence of tokens in $\mathcal{T}$ corresponding to tool observations. The masked training objective is defined as
\[
\mathcal{L}(x,\mathcal{T};\theta) \;=\; 
- \frac{\sum_{t=1}^{T} \mathbb{I}[\,y_t \notin \mathcal{O}\,] \cdot \log p_{\theta}\!\left(y_t \mid x, y_{<t}\right)}
{\sum_{t=1}^{T} \mathbb{I}[\,y_t \notin \mathcal{O}\,]},
\]
where $\mathbb{I}[\cdot]$ denotes the indicator function.  

This masking restricts the loss to the agent’s own outputs—such as internal reasoning steps and tool-calling decisions—while excluding literal tool observations. In doing so, the objective encourages the model to faithfully reproduce the supervised reasoning process (e.g., when to retrieve and how to compose intermediate steps) while ignoring raw tool outputs, a practice shown to improve both performance and robustness.

\iffalse
\subsection{Self-reflective Reasoning Chain Construction}
To construct extended reasoning chains incorporating tool interactions, we employ a multi-turn rollout strategy coupled with rejection sampling. Given a multi-hop problem instance $x_i$ from the dataset, we first prompt an initial Large Language Model (LLM) to generate a tool-augmented reasoning path $P_1$, culminating in an answer submission denoted as $\hat{y}_i$. Subsequently, a secondary LLM is prompted independently to verify the correctness of the proposed answer $\hat{y}_i$. This verification step is repeated $K$ times to produce multiple candidate verification outcomes, from which one verified sample $R_1$ aligning with factual correctness is selected. The chosen verification outcome $R_1$ is concatenated to the initial reasoning path $P_1$, forming an extended reasoning path $\{P_1, R_1\}$. The initial LLM is then employed again to refine this reasoning path, yielding $\{P_1, R_1, P_2\}$. This iterative verification and refinement process continues until either the correct answer is identified or the maximum allowable number of iterations is reached. If a correct answer emerges, the finalized reasoning path $\{P_1, R_1, P_2, \dots, R_n\}$ is recorded.
\fi
\subsection{Self-Reflective Reinforcement Learning (SRRL)}

% \yz{rewrite}

Unlike other previous training frameworks, our reinforcement learning framework unifies SFT and RL under the self-reflection mechanism, which we call Self-Reflective Reinforcement Learning (SRRL). Specifically, SRRL as allows the model to submit answers multiple times within a single dialogue turn. This design enables the model to iteratively refine its reasoning based on external feedback, leading to more stable and effective exploration.

Formally, given an input query $x$ and the ground-truth answer $y^\ast$, the LLM interacts with external tools to produce an evolving trajectory
\[
\mathcal{T} = \{(a_1,o_1), (a_2,o_2), \dots, (a_T,o_T)\},
\]
where at step $t$, $a_t$ denotes the agent’s action (e.g., a tool call or an \textit{answer\_submit}), and $o_t$ denotes the resulting observation.  
In particular, when the action corresponds to the special tool \textit{answer\_submit}, the submitted answer $\hat{y}^{(t)}$ is compared with $y^\ast$ and return
\[
r^{(t)} = \mathrm{F}_1\!\big(\hat{y}^{(t)},\,y^\ast\big) \in [0,1].
\]
The scalar feedback $r^{(t)}$ is returned \emph{as text} and appended to the dialogue context.  
If $r^{(t)}$ is below a predefined threshold, the environment allows the model to continue reasoning, enabling the model to revise its reasoning and potentially submit an improved answer at a later step.

We employ the Group Relative Policy Optimization (GRPO) algorithm~\citep{shao2024deepseekmath} to optimize the policy $\pi_\theta$ parameterized by the model. In addition, following DAPO~\citep{yu2025dapoopensourcellmreinforcement}, we incorporate the \emph{clip-higher} mechanism to better accommodate asymmetric update ranges. Concretely, given a batch of queries $(q,a) \sim \mathcal{D}$ and $T$ sampled reasoning trajectories $\{o_i\}_{i=1}^T$ from the old policy $\pi_{\theta_{\text{old}}}$, the DAPO objective is defined as

{\small
\begin{equation}
\begin{aligned}
\mathcal{J}_{\text{DAPO}}(\theta) = 
\mathbb{E}_{(q,a)\sim\mathcal{D},\, \{o_i\}\sim \pi_{\theta_{\text{old}}}(\cdot|q)}
\Bigg[
\frac{
\sum_{i=1}^G \sum_{t=1}^{|o_i|}
\min\!\Big(
r_{i,t}(\theta)\hat{A}_{i,t},\,
\text{clip}\!\big(r_{i,t}(\theta),1-\epsilon_{\text{low}},1+\epsilon_{\text{high}}\big)\hat{A}_{i,t}
\Big)
}{
\sum_{i=1}^G |o_i|
}
\Bigg],
\end{aligned}
\end{equation}
}
where $r_{i,t}(\theta)$ denotes the importance sampling ratio between the new and old policies at timestep $t$, and $\hat{A}_{i,t}$ is the estimated advantage. The asymmetric clipping parameters $\epsilon_{\text{low}}$ and $\epsilon_{\text{high}}$ control the update range, preventing overfitting to noisy advantages.

%This formulation provides robust gradient estimates across groups of trajectories, yielding more stable, diverse, and reward-sensitive learning in multi-step reasoning tasks for retrieval-augmented generation (RAG) systems.

\iffalse
\subsection{Reward Design}
Our reward function comprises two components: a format-based reward and an answer-correctness reward.
To penalize overly long responses, we adopt a length-sensitive reward inspired by DAPO. The format reward $R_{\text{length}}$ is defined as:

{\small
\[
R_{\text{length}}=
\begin{cases}
0, & \lvert y\rvert \le L_{\text{expect}},\\
\dfrac{L_{\text{expect}} - \lvert y\rvert}{L_{\text{cache}}}, 
& L_{\text{expect}} < \lvert y\rvert \le L_{\max},\\
-1, & L_{\max} < \lvert y\rvert,
\end{cases}
\]
}

where $\lvert y\rvert$ denotes the output length, $L_{\max}$ is the maximum allowed length, and $L_{\max} - L_{\text{expect}}$ defines the linear decay interval. No penalty is applied when the length is within the safe zone $L_{\text{expect}}$. In the transition band, the penalty grows linearly, and once the length exceeds $L_{\max}$, the reward saturates at $-1$.

The correctness of the model’s response is rewarded using an exponentially decaying function:

$$
R_{\text{correct}} = r \cdot \alpha^T,
$$

where $T$ is the number of submissions made in the sample, $r$ is a quality metric (we use the F\textsubscript{1} score), and $\alpha \in (0,1]$ is a discount factor that discourages repeated attempts.
\fi

\subsection{Reward Design}

Because in addition to the correctness of the answer, we also care about how many times the agent has tried, we define a trajectory-wise reward. Let a trajectory be denoted by 
$\tau = \{(o_t, a_t, r_t)\}_{t=1}^N$, where $o_t$ is the observation (context), $a_t$ is the model output 
including potential tool invocations, and $r_t$ is the instantaneous reward. The total trajectory-wise reward 
is then given by
\begin{equation}
R(\tau) = R_{\text{format}}(\tau) + R_{\text{correct}}(\tau).
\end{equation}

For format, let $\lvert y \rvert$ be the output length, $L_{\text{expect}}$ the safe-zone threshold, and $L_{\max}$ the hard limit. The reward is
\begin{equation} R_{\text{format}}(\tau) = \begin{cases} 0, & \lvert y \rvert \leq L_{\text{expect}}, \\[6pt] -\dfrac{\lvert y \rvert - L_{\text{expect}}}{L_{\max} - L_{\text{expect}}}, & L_{\text{expect}} < \lvert y \rvert \leq L_{\max}, \\[12pt] -1, & \lvert y \rvert > L_{\max}. \end{cases} \end{equation}
Thus, outputs in the safe zone incur no penalty, those in the transition region are linearly penalized, and overly long ones receive the maximum penalty.

For correctness, let $r \in [0,1]$ be a task-specific score (e.g., F\textsubscript{1}) and $T$ the number of submission attempts. To discourage resubmissions, we apply an exponential discount $\alpha \in (0,1]$:
\begin{equation}
R_{\text{correct}}(\tau) = r \cdot \alpha^T.
\end{equation}

\section{Experiments}
\subsection{Experiment Setup}

\paragraph{Datasets.}
We evaluate our model on a diverse suite of open-domain QA benchmarks, spanning both in-domain and out-of-domain settings. Following \citet{zheng2025deepresearcherscalingdeepresearch}, we adopt the same evaluation split: 512 examples sampled from the development sets of NQ~\citep{kwiatkowski-etal-2019-natural}, TQ~\citep{joshi2017triviaqalargescaledistantly}, HotpotQA~\citep{yang2018hotpotqa}, 2WikiMultiHopQA~\citep{ho2020constructingmultihopqadataset}, MuSiQue~\citep{trivedi2022musique}, PopQA~\citep{mallen-etal-2023-trust}, FanoutQA~\citep{zhu-etal-2024-fanoutqa}, FRAMES~\citep{el-asri-etal-2017-frames}, and SimpleQA~\citep{wei2024measuringshortformfactualitylarge}, along with 125 examples from Bamboogle. 

Because valid answers in open-domain QA often admit multiple surface forms, rule-based string-matching metrics can lead to inaccurate performance estimates. To address this, we adopt LLM-as-a-Judge, following the methodology and prompt template of \citet{zheng2025deepresearcherscalingdeepresearch}. During evaluation, all models are restricted to submitting a single answer. The full evaluation prompt and implementation details are provided in Appendix~\ref{app:evaluation}.

\paragraph{Baselines.}  
We compare our approach against several strong baselines that represent different paradigms for reasoning and retrieval in open-domain QA:  
(1) \textbf{Closed-book (CoT)}: The model answers questions using only its internal parametric knowledge, without any external retrieval, following a chain-of-thought prompting strategy. We choose Qwen2.5-7B-Instruct~\citep{qwen2025qwen25technicalreport} as the base model.
(2) \textbf{Local RAG}: The model is allowed to access a local vector-based retrieval system, where the knowledge source consists of the English Wikipedia page dump dated March 1, 2022. We test Qwen2.5-7B-Instruct~\citep{qwen2025qwen25technicalreport}, Search-r1~\citep{jin2025searchr1trainingllmsreason}, and R1-Searcher~\citep{song2025r1searcherincentivizingsearchcapability} in this setting.
(3) \textbf{Web Agents}: The model is equipped with web-based tool access, including search engine querying and webpage parsing through Markdown conversion. We test DeepResearcher~\citep{zheng2025deepresearcherscalingdeepresearch} in this setting.

\paragraph{Implementation Details}  
During inference, we use the Google Web Search API for real-time retrieval and the Jina API to bypass anti-crawling and extract clean, LLM-friendly text.

For training, we prioritize cost, stability, and consistency by restricting retrieval to Wikipedia via the Google Site Search API and fetching full pages through the official Wikipedia API. Training uses the verl framework~\citep{sheng2025hybridflow}, sampling 12 prompts per step, each with 8 candidate trajectories and up to 30 interaction turns. This controlled setup ensures stable, noise-reduced signals while still exposing the model to realistic retrieval and comprehension tasks. We trained a total of 100 steps, spending 60 A800 GPU hours.

\begin{table*}[!ht]
\centering
\vspace{-2mm}
\caption{Main results on seven multi-hop question answering (MHQA) benchmarks. All the results labelled with $^\dagger$ are taken from \citep{zheng2025deepresearcherscalingdeepresearch}.}
\resizebox{\textwidth}{!}
{%
\begin{tabular}{lcccccccccc}
\toprule
\multirow{2}{*}{\textbf{Method}} & \textbf{Inference} & \multicolumn{5}{c}{\textbf{In Domain}} & \multicolumn{4}{c}{\textbf{Out of Domain}} \\
& \textbf{Environment} & \textbf{NQ} & \textbf{TQ} & \textbf{Hotpot} & \textbf{2Wiki} & \textbf{Avg} & \textbf{Musique} & \textbf{Bamb} & \textbf{PopQA} & \textbf{Avg} \\
\midrule
CoT$^\dagger$ & - & 32.0 & 48.2 & 27.9 & 27.3 & 33.9 & 7.4 & 21.6 & 15.0 & 14.7 \\
CoT+RAG$^\dagger$ & Local RAG & 59.6 & 75.8 & 43.8 & 24.8 & 51.0 & 10.0 & 27.2 & 48.8 & 28.7 \\
Search-o1$^\dagger$ & Web Search & 55.1 & 69.5 & 42.4 & 37.7 & 51.2 & 19.7 & 53.6 & 43.4 & 38.9 \\
\midrule
\multicolumn{11}{l}{\textbf{\textit{7B/8B Models}}} \\
Qwen3-8B w/ Tools & Local RAG & 67.0  &  76.4 & 50.8 & 33.0  & 56.8 & 18.4  & 43.2  & 44.0  & 35.2  \\
Search-r1-base$^\dagger$ & Local RAG & 60.0 & 76.2 & 63.0 & 47.9 & 61.8 & 27.5 & 57.6 & 47.0 & 44.0 \\
Search-r1-instruct$^\dagger$ & Local RAG & 49.6 & 49.2 & 52.5 & 48.8 & 50.0 & 28.3 & 47.2 & 44.5 & 49.5 \\
R1-Searcher$^\dagger$ & Web Search & 52.3 & 79.1 & 53.1 & 65.8 & 62.6 & 25.6 & 65.6 & 43.4 & 44.9 \\
DeepResearcher$^\dagger$ & Web Search & 61.9 & 85.0 & 64.3 & 66.6 & 69.5 & 29.3 & 72.8 & 52.7 & 51.6 \\
%EvolveSearch$^\dagger$ & Web Search & \textbf{71.0} & \textbf{89.5} & \textbf{67.7} & 76.4 & \textbf{76.2} & \textbf{33.8} & \textbf{77.1} & 50.3 & \textbf{53.7} \\
\midrule
\multicolumn{11}{l}{\textbf{\textit{14B Models}}} \\
Qwen2.5-14B w/ Tools & Local RAG & 72.1 & 83.8 & 62.9 & 70.9 & 72.4 & 29.7 & 72.0 & 46.1 & 49.3 \\
Qwen2.5-14B w/ Tools & Web Search & 72.5 & 87.9 & 67.9 & 80.3 & 77.2 & 26.6 & 73.6 & 54.7 & 51.6 \\
Qwen3-14B w/ Tools & Local RAG & 73.1  & 80.9  & 54.9 & 52.5  & 65.4 & 22.7  & 63.2  & 46.7  & 44.2 \\
Search-r1 & Local RAG & 66.9 & 82.6 & 69.8 & 57.0 & 69.1 & \textbf{36.9} & 64.8 & 56.3 & 52.7 \\
WebSeer & Local RAG & 81.9 & 86.7 & 70.9 & 76.0 & 78.9 & 35.0 & \textbf{81.6} & \textbf{60.6} & \textbf{59.1} \\
WebSeer & Web Search & \textbf{82.8} & \textbf{91.0} & \textbf{72.3} & \textbf{84.2} & \textbf{82.6} & 35.2 & 80.0 & 58.0 & 57.7 \\
\bottomrule
\end{tabular}%
}
\vspace{-2mm}
\label{tab:mainresult}
\end{table*}

\subsection{Main Results}
 \begin{wrapfigure}{r}{0.5\textwidth}
\vspace{-8pt} % 可以根据需要调节上下间距
\centering
\captionsetup{type=table}
\caption{Evaluation on three harder benchmarks. Qwen2.5-14B and WebSeer use web search engine, while Search-r1 relies on local RAG.}
\resizebox{\linewidth}{!}{%
\begin{tabular}{@{}lcccc@{}}
\toprule
Model& FanoutQA & FRAMES & SimpleQA & Avg. \\  \midrule
Qwen2.5-14B & 45.5       &  52.7      &   85.7      &  61.3   \\
Search-r1-14B  & 12.6       &  29.5      &   36.4      &  26.2  \\
WebSeer   & \textbf{55.4}       &  \textbf{56.1}      &   \textbf{90.0}      &  \textbf{65.3}  \\ 
\bottomrule
\end{tabular}
}
\label{tab:res2}
\vspace{-8pt} % 避免和正文间隔太大
\end{wrapfigure}

Table~\ref{tab:mainresult} reports the performance of our method against baselines on seven multi-hop QA benchmarks. Our approach consistently achieves the best results, substantially outperforming both closed-book and retrieval-augmented baselines. On in-domain tasks, our model reaches an average accuracy of 82.4\%, exceeding the previous state-of-the-art method Search-r1 by 12.5 points. The largest gains are observed on NQ and 2WikiMultiHopQA, with improvements of 15.9 and 27.2 points, respectively.

\begin{wrapfigure}{r}{0.55\textwidth}
\vspace{-8pt} % 可调，避免和上文间隔太大
\centering
\captionsetup{type=table}
\caption{Accuracy and average tool call times for Qwen2.5 Models on HotpotQA and SimpleQA}
\setlength{\tabcolsep}{2pt} % 控制列间距
\resizebox{\linewidth}{!}{%
\begin{tabular}{@{}p{1.2cm}cccc@{}}
\toprule
\textbf{Model} & \multicolumn{2}{c}{\textbf{HotpotQA}} & \multicolumn{2}{c}{\textbf{SimpleQA}} \\
\cmidrule(lr){2-3} \cmidrule(lr){4-5}
  & Acc & Tool Call & Acc & Tool Call \\
\midrule
\multicolumn{5}{@{}l}{\textbf{Qwen2.5-3B}} \\
Instruct & 44.73 & 4.31  & 41.02 & 4.17 \\
SFT      & 41.21 \textcolor{red}{(-3.52)} & 12.40 & 49.08 \textcolor{deepgreen}{(+8.06)} & 11.46 \\
\midrule
\multicolumn{5}{@{}l}{\textbf{Qwen2.5-7B}} \\
Instruct & 51.95 & 2.95  & 51.56 & 3.24 \\
SFT      & 46.09 \textcolor{red}{(-5.86)} & 9.23  & 50.39 \textcolor{red}{(-1.17)} & 11.09 \\
\midrule
\multicolumn{5}{@{}l}{\textbf{Qwen2.5-14B}} \\
Instruct & 62.89 & 3.57  & 65.43 & 3.76 \\
SFT      & 68.75 \textcolor{deepgreen}{(+5.86)} & 13.43 & 76.17 \textcolor{deepgreen}{(+10.74)} & 10.82 \\
\bottomrule
\end{tabular}%
}
\label{tab:ab1}
\vspace{-8pt} % 避免和下文间隔太大
\end{wrapfigure}

Beyond in-domain evaluation, our method also demonstrates strong generalization on out-of-distribution (OOD) datasets, indicating that it does not merely overfit to the retrieval distribution encountered during training. Instead, it learns reasoning patterns and retrieval strategies that transfer effectively to unseen question types, domains, and web sources. In this regime, WebSeer benefits from local RAG: on Bamboogle, it achieves 81.6\%, a substantial 12.8-point improvement over the prior best, while on PopQA it reaches 60.6\%. These results highlight the effectiveness of our reinforcement learning framework and tool-augmented reasoning design in enabling robust cross-domain generalization.

We further evaluate on three challenging benchmarks: FanOutQA, Frames, and SimpleQA (Table~\ref{tab:res2}). On FanOutQA, a fully OOD multi-document QA benchmark, our model attains a loose accuracy of 55.4, surpassing all baselines and nearly matching GPT-4o (55.8)\footnote{\url{https://fanoutqa.com/leaderboard/}}. It also achieves 56.1\% on Frames and 90.0\% on SimpleQA, while the RL-only Search-r1 model performs poorly across these datasets. These results underscore the strong generalization of our approach: despite being trained under site-restricted search, it performs even better when deployed in the open web, demonstrating a robust and transferable retrieval–reasoning policy that adapts to diverse domains.

\subsection{Quantitative Analysis}

\paragraph{Model capacity matters for complex tool using.}  
We find that sufficient model capacity is essential for multi-step reasoning in search agents. As shown in Table~\ref{tab:ab1}, SFT consistently increases tool usage across scales but its effect on accuracy is uneven: the 3B model drops 3.52 points on HotpotQA yet gains 8.06 on SimpleQA, while the 7B model degrades on both. Only the 14B model achieves consistent improvements in both tool usage and accuracy, underscoring the role of scale.  

Applying RL after SFT reinforces this pattern: the 14B model improves steadily, while smaller models (3B, 7B) show little benefit and suffer from instability, including repetitive text and malformed JSON that causes failed tool calls. Although rewards may rise initially, behavior often collapses. Overall, sufficient scale is crucial for stable reasoning and reliable tool-augmented decision making.

\paragraph{Training progressively shapes tool-use behavior from underuse to strategic deployment.}
\begin{wrapfigure}{r}{0.5\textwidth}
\vspace{-18pt}
\centering
\includegraphics[width=\linewidth]{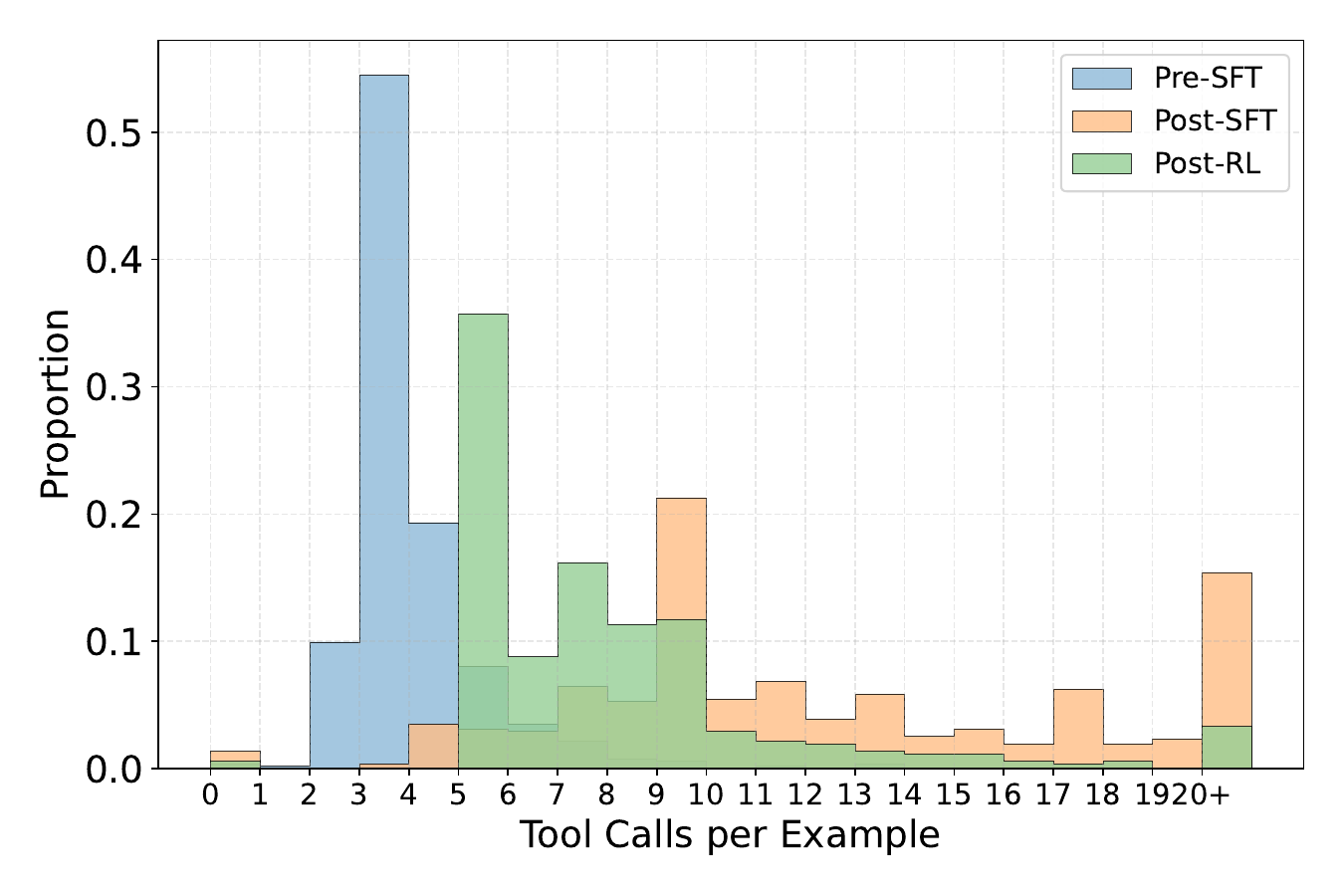}
\vspace{-3mm}
\caption{Tool call distributions on HotpotQA across three training stages: Pre-SFT, Post-SFT, and Post-RL.}
\label{fig:tcd}
\vspace{-13pt}
\end{wrapfigure}

We examine tool invocation distributions across three development stages: pre-SFT, post-SFT, and post-RL. Using HotpotQA trajectories, we plot interaction counts per example (Figure~\ref{fig:tcd}), revealing how supervision and reinforcement learning shape reasoning depth and tool use. Before SFT, tool usage is limited, with most conversations involving around three calls—suggesting a conservative strategy arising from insufficient mastery of tool behaviors. After SFT, the distribution shifts markedly rightward, peaking at 10 calls and extending up to 50, indicating more active and flexible tool engagement, often in lengthy multi-step interactions. Following RL, the distribution sharpens between 5 and 8 calls, with very high and very low counts becoming rare.

Although we did not penalize underuse during RL, the model rarely produces trajectories with fewer than five calls. This implies that repeated tool use is implicitly reinforced, as it aids verification and validation. Overall, training progression shows a shift from underuse, to overuse, to strategic use. RL fine-tuning improves task performance and yields more stable, efficient behavior, encouraging sufficient—but not excessive—tool invocation without hard-coded constraints.
\begin{wrapfigure}{r}{0.5\textwidth}
\vspace{-8pt} % 调整与上文距离
\centering
\includegraphics[width=\linewidth]{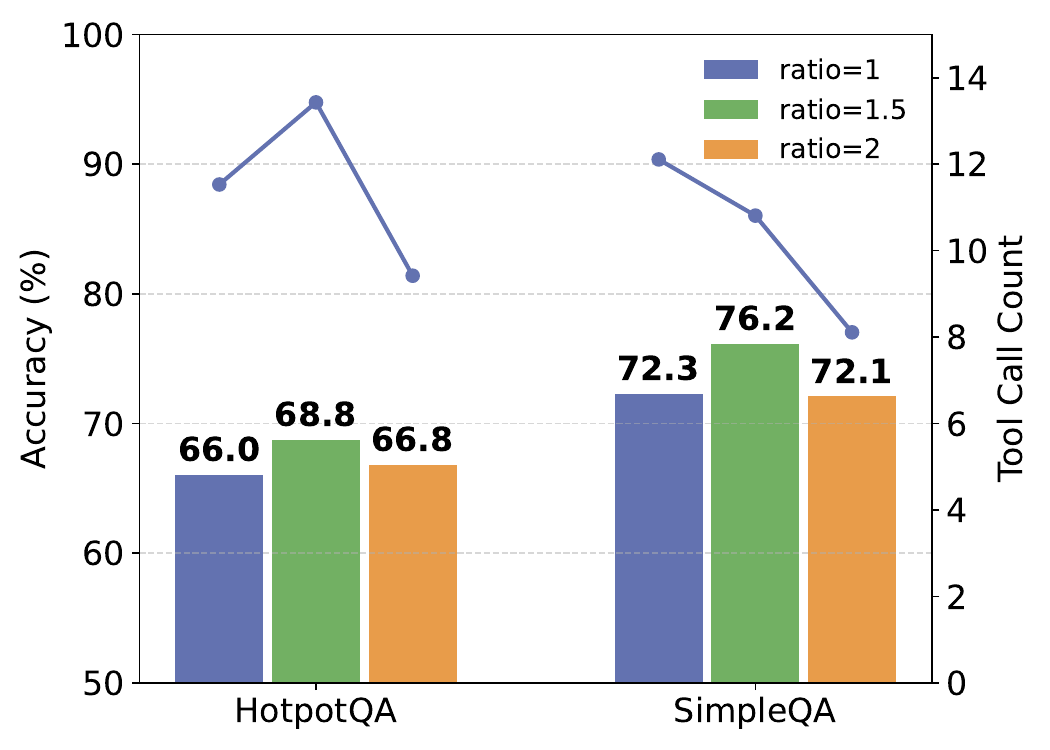}
\caption{Impact of SFT data composition on tool usage and accuracy. 
We vary the ratio between single-pass correct trajectories and multi-refinement trajectories during SFT.}
\label{fig:f1}
\vspace{-8pt} % 调整与下文距离
\end{wrapfigure}

\textbf{Data mixing ratio in SFT are also key to performance improvement.}

We further examine the effect of data composition during the SFT stage. As shown in Figure~\ref{fig:f1}, the ratio between single-pass correct trajectories—where the model produces the correct answer in a single step—and multi-refinement trajectories—where multiple reasoning or retrieval steps are required—plays a pivotal role in shaping model behavior. Increasing the share of longer reasoning trajectories encourages more frequent tool usage, but does not necessarily translate into higher accuracy. Striking an appropriate balance between the two types of trajectories is therefore essential for effective SFT, and can even determine the success of subsequent RL fine-tuning.

\subsection{Ablation Studies}

\begin{wrapfigure}{r}{0.55\textwidth}
\vspace{-8pt} % 上方间距可调
\centering

% 上半部分表格
\begin{minipage}{\linewidth}
\centering
\setlength{\tabcolsep}{2pt} % 缩小列间距
\begin{tabular}{lcccc}
\toprule
& \multicolumn{2}{c}{HotpotQA} & \multicolumn{2}{c}{SimpleQA} \\
\cmidrule(r){2-3} \cmidrule(r){4-5}
Method & Acc & Tool Call & Acc & Tool Call \\
\midrule
WebSeer & 70.90 & 7.91 & 78.91 & 8.61  \\
\quad \small{- \textit{w/o SRRL}}  & 67.27 & 7.38 & 75.98 & 6.15  \\
\bottomrule
\end{tabular}
\end{minipage}

\vspace{0.2cm} % 表格与图片间距

% 下半部分图片
\begin{minipage}{\linewidth}
\centering
\includegraphics[width=\linewidth]{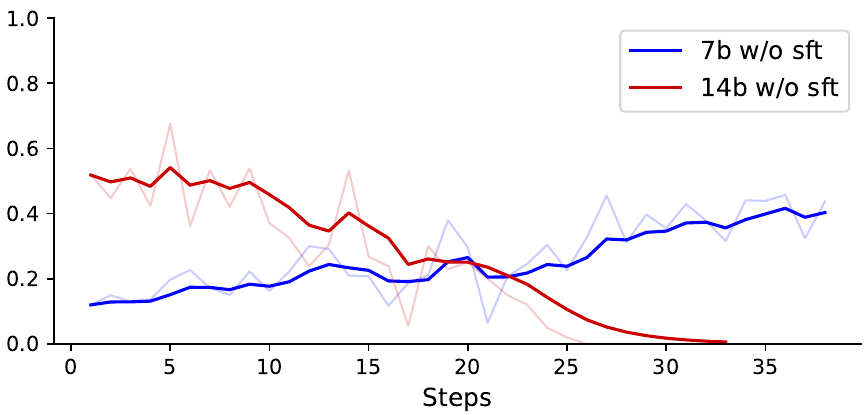} % 自动适应 wrap 宽度
\end{minipage}

\caption{
(Top) Limiting the model to submit only one answer results in a decrease in the final model's performance. 
(Bottom) Without complex data for SFT, the 14B model collapses with decreasing rewards, while the 7B model shows slight reward improvement but often generates garbled text before tool calls, potentially leading to failure.
}
\label{fig:combined}
\vspace{-48pt} % 下方间距可调
\end{wrapfigure}

In this section, we present ablation experiments to assess the contribution of key components in our framework. We evaluate two variants: (1) restricting the model to produce only a single answer during reinforcement learning, and (2) training without cold-start initialization. Results are summarized in Table~\ref{fig:combined}.

Our analysis yields two main observations:  
(1) Each component is critical to the success of training—removing any of them consistently degrades performance.  
(2) The cold-start strategy is especially important, as it substantially improves the model’s ability to develop effective tool-use behaviors. For more challenging tasks, we further find that high-quality SFT data is indispensable for stable training.

\section{Related Work}
\iffalse
\textbf{Large Reasoning Models.}
Early efforts to enhance the reasoning capabilities of large language models (LLMs) primarily relied on prompting techniques that encouraged step-by-step reasoning, such as Chain-of-Thought (CoT)~\citep{wei2023chainofthoughtpromptingelicitsreasoning}, Tree-of-Thought (ToT)~\citep{yao2023treethoughtsdeliberateproblem}, and Least-to-Most~\citep{zhou2023leasttomostpromptingenablescomplex} prompting. While effective to some extent, these approaches often generate relatively rigid and static reasoning trajectories. More recently, models such as OpenAI’s O1~\citep{openai2024openaio1card} and DeepSeek’s R1~\citep{deepseekai2025deepseekr1incentivizingreasoningcapability} have adopted reinforcement learning-based training within dynamic and interactive environments. This paradigm enables models to develop more flexible, context-sensitive, and extended chains of thought, thereby significantly improving performance on complex reasoning tasks.
However, these models exhibit more hallucinations in question-answering tasks and do not integrate tool calls into the reasoning process.
\fi
\textbf{LLM With Tools}
A growing body of work focuses on tool-augmented LLMs which models designed to interact with external tools (e.g., APIs, databases) to perform complex tasks. ToolLLaMA~\citep{qin2023toolllmfacilitatinglargelanguage} applies structured decision-tree search to decompose multi-step instructions into tool calls, demonstrating that explicit planning over API chains enhances compositional reasoning. The ToolPlanner~\citep{wu2024toolplannertoolaugmentedllm} model further refines this dynamic, shifting from overly detailed instruction prompting to more realistic interaction schemas across multi-granularity instruction tasks. However, limitations remain in completing harder tasks. Tool-Star~\citep{dong2025toolstarempoweringllmbrainedmultitool} generates tool usage trajectories as cold-start data through prompt-based sampling, then designs a multi-tool self-critique reinforcement learning framework with hierarchical reward design to enhance the LLM's reward understanding in real-world scenarios and promote effective tool collaboration, representing the latest training paradigm.

\textbf{Reasoning Agentic RAG} 
Early RAG approaches are primarily linear or branching structures~\citep{Chen_Lin_Han_Sun_2024, gao2024retrievalaugmentedgenerationlargelanguage, gao2024modularragtransformingrag}. They typically rely on manually crafted prompts or fixed execution workflows, which severely constrain the model’s autonomy and flexibility. Recent work has begun to incorporate reinforcement learning into RAG~\citep{jin2025searchr1trainingllmsreason,song2025r1searcherincentivizingsearchcapability}. These methods adopt an outcome-driven RL framework that enables the model to explore how to invoke external retrieval systems during the reasoning process. 
DeepResearcher~\citep{zheng2025deepresearcherscalingdeepresearch} extends this line of work to the web search setting, but their approach still depends on stronger models to act as agents for webpage navigation. \citep{shi2025pangudeepdiveradaptivesearch} propose Pangu DeepDiver, which combines a carefully constructed dataset designed to foster information-seeking behavior in open-world internet environments with a specialized DeepDiver framework to enhance search capabilities.
However, despite these advances, the reasoning chains produced by current methods remain relatively shallow and are insufficient for solving more complex or open-ended tasks.

\section{Conclusion}
In this work, we introduced WebSeer, a novel agent training paradigm tailored for real-world web-based retrieval environments. By synthesizing multi-refinement reasoning trajectories through rejection sampling and incorporating self-reflective reinforcement learning (SRRL), WebSeer learns to perform deeper, more robust reasoning that mimics human information-seeking behavior. Through extensive experiments across a wide range of open-domain and out-of-domain question answering benchmarks, WebSeer consistently outperforms existing baselines achieving state-of-the-art performance on datasets such as HotpotQA, Bamboogle, and SimpleQA. 
% Our ablations further confirm that model scale, training stages, and high-quality reasoning data all play critical roles in shaping effective tool-use policies. 
WebSeer lays a foundation for more general-purpose reasoning agents that can seamlessly interact with dynamic, heterogeneous web environments.

\clearpage

\bibliography{iclr2026_conference}
\bibliographystyle{iclr2026_conference}

\appendix
\section{Appendix}
\label{app:evaluation}
\begin{tcolorbox}[breakable,title=Prompts for WebSeer]
System: \\
You are a reasoning assistant with the ability to perform web searches and execute Python code to help you process the content of the page and answer the user's question accurately. \\ 
Do not use any knowledge you know; all facts in your thinking should be obtained from the information returned by the tools. You can repeat the search process multiple times if necessary.  \\ 
Once you have all the information you need, continue your reasoning. \\ 
Please first make a plan before calling tools. \\ 
Please answer the following question. You should provide your final answer to the "submit\_answer" tool. \\ 

Tools:

You may call one or more functions to assist with the user query.

You are provided with function signatures within <tools></tools> XML tags: \\ 
<tools> \\ 
{"type": "function", "function": {"name": "submit\_answer", "description": "Submit your final answer. You must use this tool to submit your answer before the dialog ends.", "parameters": {"type": "object", "properties": {"answer": {"type": "string", "description": "Your final answer"}}, "required": ["answer"]}}} \\ 
{"type": "function", "function": {"name": "search", "description": "Call google to search for relevant information.", "parameters": {"type": "object", "properties": {"query": {"type": "string", "description": "Search keywords"}}, "required": ["query"]}}} \\ 
{"type": "function", "function": {"name": "query\_on\_page", "description": "This tool will visit a specific page of url, and it will answer the question based on the content of the page. The assistant has no context information, please describe the question completely.", "parameters": {"type": "object", "properties": {"url": {"type": "string", "description": "The url of the page, must be a page provided by the search tool."}, "question": {"type": "string", "description": "The question about the content of the page"}}, "required": ["url", "question"]}}} \\ 
</tools> \\ 

For each function call, return a json object with function name and arguments within <tool\_call></tool\_call> XML tags: \\ 
<tool\_call> \\ 
{"name": <function-name>, "arguments": <args-json-object>} \\ 
</tool\_call> \\
user: \\
Question: {Question}

\end{tcolorbox}

\begin{tcolorbox}[breakable,title=Prompts for Evaluation]
You will be given a question and its ground truth answer list where each item can be a ground truth answer. Provided a pred\_answer, you need to judge if the pred\_answer correctly answers the question based on the ground truth answer list. \\
You should first give your rationale for the judgement, and then give your judgement result (i.e., correct or incorrect). \\

Here is the criteria for the judgement: \\
1. The pred\_answer doesn't need to be exactly the same as any of the ground truth answers, but should be semantically same for the question. \\
2. Each item in the ground truth answer list can be viewed as a ground truth answer for the question, and the pred\_answer should be semantically same to at least one of them. \\

question: \{question\} \\
ground truth answers: \{target\} \\
pred\_answer: \{predicted\_answer\} \\

The output should in the following json format: \\

\{ \\
"rationale": "your rationale for the judgement, as a text", \\
"judgement": "your judgement result, can only be 'correct' or 'incorrect'" \\
\} \\

Your output: \\

\end{tcolorbox}

\section{Case study}

Case study 1.1 is generated by the Qwen2.5-14B-instruct model and exhibits a brittle reasoning process; it over-relies on information parsed from search snippets, leading to an incorrect inference about one player's batting hand, and prematurely abandons a sub-task after a single, poorly formulated query fails. This results in an incomplete and partially erroneous answer. In contrast, WebSeer employs a more robust, iterative strategy. It consistently pairs web searches with targeted page queries for verification, avoiding the errors from snippet parsing. Crucially, when faced with an initial search failure for a specific entity (player Jeff Austin), it demonstrates effective error recovery by refining its search query with disambiguating context (`1998 MLB draft`). This methodical approach of iterative refinement and verification, despite requiring more tool interactions, allows the second agent to successfully decompose the problem and provide a complete and correct solution.

Case study 2.1 involved more than a dozen tool calls: multiple search calls to locate the “Panther tank” article and repeated query\_on\_page checks to confirm it was officially the Panzerkampfwagen V Panther deployed on both Eastern and Western fronts from mid-1943, followed by additional query\_on\_page calls verifying that the Jagdpanther tank destroyer used its armor and suspension; by repeatedly cross-validating the tank’s name, deployment timeline, and derivative chassis, the process covered all key details and reinforced the conclusion with multiple lines of evidence, perfectly demonstrating how to leverage tools systematically and efficiently to support rigorous reasoning.
\label{app:cs1}
\begin{tcolorbox}[breakable, fontupper=\small,title=Case study 1.1 (Qwen2.5-14B-instruct)]
\textbf{System:}

You are a reasoning assistant with the ability to perform web searches and execute Python code to help you process the content of the page and answer the user's question accurately.
Do not use any knowledge you know; all facts in your thinking should be obtained from the information returned by the tools. You can repeat the search process multiple times if necessary. 
Once you have all the information you need, continue your reasoning.
Please first make a plan before calling tools.
Please answer the following question. You should provide your final answer to the "submit\_answer" tool.

\vspace{1em}

\textbf{User:}

Question: What is the batting hand of each of the first five picks in the 1998 MLB draft?
\vspace{1em}

\textbf{Assistant:}

To find the batting hand of each of the first five picks in the 1998 MLB draft, I will follow these steps:

1. Search for the list of the first five picks in the 1998 MLB draft.
2. For each player, search for their batting hand (left-handed or right-handed).

Let's start with step 1.

Tool Calls:
search(query=first five picks in the 1998 MLB draft)
\vspace{1em}

\textbf{Tool Response (search):}

\begin{lstlisting}
1. Title: 1998 Major League Baseball draft - Wikipedia
URL: https://en.wikipedia.org/wiki/1998_Major_League_Baseball_draft
Content Summary:
<b>First</b> round <b>selections</b> ; <b>5</b>, J. D. Drew &middot; St. Louis Cardinals ; 6, Ryan Mills, Minnesota Twins ; 7, Austin Kearns &middot; Cincinnati Reds ; 8, Felipe Lopez &middot; Toronto Blue&nbsp;...

2. Title: List of 1998 Seattle Mariners draft picks - Wikipedia
URL: https://en.wikipedia.org/wiki/List_of_1998_Seattle_Mariners_draft_picks
Content Summary:
 In all, the Mariners selected 23 pitchers, 9 outfielders, 7 catchers, <b>5</b> shortstops, 3 <b>first</b> basemen, 2 third basemen, and 1 second baseman.

3. Title: List of Major League Baseball first-round draft picks - Wikipedia
URL: https://en.wikipedia.org/.../List_of_Major_League_Baseball_first-round_dr...
Content Summary:
 Contents &middot; 1 Atlanta Braves; 1.1. &middot; 2 Miami Marlins; 1.1. &middot; 3 New York Mets; 1.1. &middot; 4 Philadelphia Phillies; 1.1. &middot; <b>5</b> Washington Nationals.
... (content truncated to save space)
\end{lstlisting}
\vspace{0.5em}\hrule\vspace{0.5em}

\textbf{Assistant:}
From the search results, the most relevant link appears to be the Wikipedia page titled "1998 Major League Baseball draft". This page contains the list of the first five picks in the 1998 MLB draft. Let's extract that information from the page.

Next, I'll visit the page and gather the names of the first five picks.

Tool Calls:
query\_on\_page(url=https://en.wikipedia.org/wiki/1998\_Major\_League\_Baseball\_draft, question=What are the names of the first five picks in the 1998 MLB draft?)
\vspace{1em}

\textbf{Tool Response (query\_on\_page):}

\begin{lstlisting}
The first five picks in the 1998 MLB draft are as follows:

1. **Pat Burrell** - Philadelphia Phillies
2. **Mark Mulder** - Oakland Athletics
3. **Corey Patterson** - Chicago Cubs
4. **Jeff Austin** - Kansas City Royals
5. **J. D. Drew** - St. Louis Cardinals
\end{lstlisting}
\vspace{0.5em}\hrule\vspace{0.5em}

\textbf{Assistant:}

Now we have the names of the first five picks in the 1998 MLB draft. Next, I will search for the batting hand of each player.

I will start with Pat Burrell.

Tool Calls:
search(query=Pat Burrell batting hand)
\vspace{1em}

\textbf{Tool Response (search):}

\begin{lstlisting}
1. Title: Pat Burrell - Wikipedia
URL: https://en.wikipedia.org/wiki/Pat_Burrell
Content Summary:
 <b>Patrick</b> Brian <b>Burrell</b> (born October 10, 1976), nicknamed &quot;<b>Pat</b> the <b>Bat</b>&quot;, is an American former professional <b>baseball</b> outfielder and current <b>hitting</b> coach for&nbsp;...

2. Title: Brian Wilson (baseball) - Wikipedia
URL: https://en.wikipedia.org/wiki/Brian_Wilson_(baseball)
Content Summary:
 Brian <b>Patrick</b> Wilson (born March 16, 1982), nicknamed &quot;the Beard&quot; and &quot;B-Weezy&quot;, is an American former professional <b>baseball</b> relief pitcher.

3. Title: List of Major League Baseball career assists leaders - Wikipedia
URL: https://en.wikipedia.org/.../List_of_Major_League_Baseball_career_assists_l...
Content Summary:
 The second baseman often possesses quick <b>hands</b> and feet, needs the ability to ... <b>Pat Burrell</b>, 92, 0, 92. Hank Sauer, 92, 0, 92. 46, Joe Vosmik, 90, 89, 1. 47&nbsp;...
... (content truncated to save space)
\end{lstlisting}
\vspace{0.5em}\hrule\vspace{0.5em}

\textbf{Assistant:}

From the search results, the Wikipedia page for Pat Burrell provides the most relevant information. Let's extract the batting hand of Pat Burrell from this page.

I will now visit the page and gather the required information.

Tool Calls:
query\_on\_page(url=https://en.wikipedia.org/wiki/Pat\_Burrell, question=What is Pat Burrell's batting hand?)
\vspace{1em}

\textbf{Tool Response (query\_on\_page):}

\begin{lstlisting}
Pat Burrell batted right-handed.
\end{lstlisting}
\vspace{0.5em}\hrule\vspace{0.5em}

\textbf{Assistant:}

Pat Burrell batted right-handed. Now let's find out the batting hand of the next player, Mark Mulder.

I will search for "Mark Mulder batting hand".

Tool Calls:
search(query=Mark Mulder batting hand)
\vspace{1em}

\textbf{Tool Response (search):}

\begin{lstlisting}
1. Title: Mark Mulder - Wikipedia
URL: https://en.wikipedia.org/wiki/Mark_Mulder
Content Summary:
 <b>Mark</b> Alan <b>Mulder</b> (born August 5, 1977) is an American former professional <b>baseball</b> player. A left-handed starting pitcher, <b>Mulder</b> pitched in Major League&nbsp;...

2. Title: List of Major League Baseball annual wins leaders - Wikipedia
URL: https://en.wikipedia.org/.../List_of_Major_League_Baseball_annual_wins_l...
Content Summary:
 <b>Mark Mulder</b>, 21, Oakland Athletics &middot; Roger Clemens &middot; Jamie Moyer, 20. 2002 &middot; Barry Zito, 23, oakland Athletics &middot; Derek Lowe, 21. 2003 &middot; Roy Halladay , 22&nbsp;...

3. Title: 2006 St. Louis Cardinals season - Wikipedia
URL: https://en.wikipedia.org/wiki/2006_St._Louis_Cardinals_season
Content Summary:
 Momentum would be slowed by injuries, as starting pitcher <b>Mark Mulder</b> was ... <b>Batting</b>. edit. Note: G = Games played; AB = At <b>bats</b>; R = Runs; H = Hits; 2B&nbsp;...
... (content truncated to save space)
\end{lstlisting}
\vspace{0.5em}\hrule\vspace{0.5em}

\textbf{Assistant:}

From the search results, the Wikipedia page for Mark Mulder provides the most relevant information. According to the summary, Mark Mulder is described as a left-handed starting pitcher. Therefore, we can conclude that Mark Mulder batted left-handed.

Next, I will search for the batting hand of Corey Patterson.

I will search for "Corey Patterson batting hand".

Tool Calls:
search(query=Corey Patterson batting hand)
\vspace{1em}

\textbf{Tool Response (search):}

\begin{lstlisting}
1. Title: Corey Patterson - Wikipedia
URL: https://en.wikipedia.org/wiki/Corey_Patterson
Content Summary:
 Atlanta, Georgia, U.S.. <b>Batted</b>: Left. Threw: Right. MLB debut. September 18, 2000, for the Chicago Cubs.

2. Title: All-Star Final Vote - Wikipedia
URL: https://en.wikipedia.org/wiki/All-Star_Final_Vote
Content Summary:
 On Monday July 7, 2003, Kenny Lofton was added to the ballot to replace the Chicago Cubs&#39; <b>Corey Patterson</b> who was injured on the day before. ... <b>hand</b> and wearing&nbsp;...

3. Title: Cody Ross - Wikipedia
URL: https://en.wikipedia.org/wiki/Cody_Ross
Content Summary:
 He is one of the few Major League players to <b>bat</b> right-handed and throw left-handed. Cody Ross. Ross with the San Francisco Giants in 2011. Outfielder.
... (content truncated to save space)
\end{lstlisting}
\vspace{0.5em}\hrule\vspace{0.5em}

\textbf{Assistant:}
From the search results, the Wikipedia page for Corey Patterson provides the most relevant information. According to the summary, Corey Patterson batted left-handed.

Next, I will search for the batting hand of Jeff Austin.

I will search for "Jeff Austin batting hand".

However, since Jeff Austin might not be as well-known as the other players, I will also include his full name and position in the search query to ensure accuracy.

I will search for "Jeff Austin, catcher batting hand".

Tool Calls:
search(query=Jeff Austin, catcher batting hand)
\vspace{1em}

\textbf{Tool Response (search):}

\begin{lstlisting}
1. Title: Austin Hedges - Wikipedia
URL: https://en.wikipedia.org/wiki/Austin_Hedges
Content Summary:
 <b>Austin</b> Charles Hedges (born August 18, 1992) is an American professional <b>baseball catcher</b> for the Cleveland Guardians of Major League <b>Baseball</b> (MLB).

2. Title: Austin Nola - Wikipedia
URL: https://en.wikipedia.org/wiki/Austin_Nola
Content Summary:
 <b>Austin</b> Kyle Nola (born December 28, 1989) is an American professional <b>baseball catcher</b> for the Colorado Rockies of Major League <b>Baseball</b> (MLB).

3. Title: Catcher - Wikipedia
URL: https://en.wikipedia.org/wiki/Catcher
Content Summary:
 ... <b>hitting</b>, defensive specialists were Jerry Grote, <b>Jeff</b> Mathis ... <b>hand</b> in professional <b>baseball</b> players with emphasis on digital ischemia in <b>catchers</b>&quot;.
... (content truncated to save space)
\end{lstlisting}
\vspace{0.5em}\hrule\vspace{0.5em}

\textbf{Assistant:}

None of the search results directly mention Jeff Austin's batting hand. However, the Cincinnati Reds all-time roster mentions Jeff Austin, which suggests he played for the Cincinnati Reds. To get accurate information about Jeff Austin's batting hand, I will visit the Wikipedia page for Jeff Austin and check his details.

I will now visit the page and gather the required information.

Since there isn't a direct link to Jeff Austin's Wikipedia page, I will search for it directly.

I will search for "Jeff Austin Cincinnati Reds batting hand".

But since the search did not yield a direct result, I will assume that Jeff Austin's batting hand is not widely documented online. Given the lack of specific information, I will proceed to the next player, J.D. Drew, and then submit the answers I have gathered so far.

Let's move on to J.D. Drew.

I will search for "J.D. Drew batting hand".

Tool Calls:
search(query=J.D. Drew batting hand)
\vspace{1em}

\textbf{Tool Response (search):}

\begin{lstlisting}
1. Title: J. D. Drew - Wikipedia
URL: https://en.wikipedia.org/wiki/J._D._Drew
Content Summary:
 David Jonathan &quot;<b>J. D.</b>&quot; <b>Drew</b> (born November 20, 1975) is an American former Major League <b>Baseball</b> right fielder. He began his major league career in 1998&nbsp;...

2. Title: Home run - Wikipedia
URL: https://en.wikipedia.org/wiki/Home_run
Content Summary:
 <b>J. D. Drew</b> has been part of two different sets of back-to-back-to-back-to ... &#39;Ralph,&#39; he said, &#39;the right-<b>hand batters</b> who hit to the right field are driving&nbsp;...

3. Title: Adam Wainwright - Wikipedia
URL: https://en.wikipedia.org/wiki/Adam_Wainwright
Content Summary:
 After the 2003 season, the Braves traded him to the Cardinals for outfielder <b>J. D. Drew</b>. ... <b>batting</b> practice on February 21. Three days later, the&nbsp;...
... (content truncated to save space)
\end{lstlisting}
\vspace{0.5em}\hrule\vspace{0.5em}

\textbf{Assistant:}

From the search results, the Wikipedia page for J.D. Drew provides the most relevant information. According to the summary, J.D. Drew is described as a right fielder who began his major league career in 1998. Additionally, the page states that he is a right-handed batter.

Now we have the batting hands for four out of the five players:
1. Pat Burrell - Right-handed
2. Mark Mulder - Left-handed
3. Corey Patterson - Left-handed
4. J.D. Drew - Right-handed

Since we couldn't find specific information about Jeff Austin's batting hand, I will assume that we don't have enough information to determine it. 

Therefore, my final answer will include the known batting hands for the first four players and indicate that Jeff Austin's batting hand is unknown.

I will now submit the answer.

\vspace{0.5em}\hrule\vspace{0.5em}

\textbf{Final Answer:}

Pat Burrell: Right-handed, Mark Mulder: Left-handed, Corey Patterson: Left-handed, Jeff Austin: Unknown, J.D. Drew: Right-handed

\end{tcolorbox}

\begin{tcolorbox}[breakable, fontupper=\small,title=Case study 1.2 (WebSeer-14B)]
\textbf{System:}

You are a reasoning assistant with the ability to perform web searches and execute Python code to help you process the content of the page and answer the user's question accurately.
Do not use any knowledge you know; all facts in your thinking should be obtained from the information returned by the tools. You can repeat the search process multiple times if necessary. 
Once you have all the information you need, continue your reasoning.
Please answer the following question. You should provide your final answer to the "submit\_answer" tool.

\vspace{1em}

\textbf{User:}
Question: What is the batting hand of each of the first five picks in the 1998 MLB draft?
\vspace{1em}

\textbf{Assistant:}
Tool Calls:
search(query=1998 MLB draft first five picks)
\vspace{1em}

\textbf{Tool Response (search):}

\begin{lstlisting}
1. Title: 1998 Major League Baseball draft - Wikipedia
URL: https://en.wikipedia.org/wiki/1998_Major_League_Baseball_draft
Content Summary:
 <b>First</b> round <b>selections</b> ; 2, Mark Mulder &middot; Oakland Athletics ; 3, Corey Patterson &middot; Chicago Cubs ; 4, Jeff Austin &middot; Kansas City Royals ; <b>5</b>, J. D. Drew &middot; St. Louis&nbsp;...

2. Title: List of Major League Baseball first-round draft picks - Wikipedia
URL: https://en.wikipedia.org/.../List_of_Major_League_Baseball_first-round_dr...
Content Summary:
 Contents &middot; 1 Atlanta Braves; 1.1. &middot; 2 Miami Marlins; 1.1. &middot; 3 New York Mets; 1.1. &middot; 4 Philadelphia Phillies; 1.1. &middot; <b>5</b> Washington Nationals.

3. Title: List of 1998 Seattle Mariners draft picks - Wikipedia
URL: https://en.wikipedia.org/wiki/List_of_1998_Seattle_Mariners_draft_picks
Content Summary:
 In all, the Mariners selected 23 pitchers, 9 outfielders, 7 catchers, <b>5</b> shortstops, 3 <b>first</b> basemen, 2 third basemen, and 1 second baseman.
... (content truncated to save space)
\end{lstlisting}
\vspace{0.5em}\hrule\vspace{0.5em}

\textbf{Assistant:}
Tool Calls:
query\_on\_page(url=https://en.wikipedia.org/wiki/1998\_Major\_League\_Baseball\_draft, question=What are the names of the first five picks in the 1998 MLB draft?)
\vspace{1em}

\textbf{Tool Response (query\_on\_page):}

\begin{lstlisting}
The first five picks in the 1998 MLB draft are as follows:

1. **Pat Burrell** - Philadelphia Phillies
2. **Mark Mulder** - Oakland Athletics
3. **Corey Patterson** - Chicago Cubs
4. **Jeff Austin** - Kansas City Royals
5. **J. D. Drew** - St. Louis Cardinals
\end{lstlisting}
\vspace{0.5em}\hrule\vspace{0.5em}

\textbf{Assistant:}

Tool Calls:
search(query=Pat Burrell batting hand)
\vspace{1em}

\textbf{Tool Response (search):}

\begin{lstlisting}
1. Title: Pat Burrell - Wikipedia
URL: https://en.wikipedia.org/wiki/Pat_Burrell
Content Summary:
 <b>Patrick</b> Brian <b>Burrell</b> (born October 10, 1976), nicknamed &quot;<b>Pat</b> the <b>Bat</b>&quot;, is an American former professional <b>baseball</b> outfielder and current <b>hitting</b> coach for&nbsp;...

2. Title: Brian Wilson (baseball) - Wikipedia
URL: https://en.wikipedia.org/wiki/Brian_Wilson_(baseball)
Content Summary:
 Brian <b>Patrick</b> Wilson (born March 16, 1982), nicknamed &quot;the Beard&quot; and &quot;B-Weezy&quot;, is an American former professional <b>baseball</b> relief pitcher.

3. Title: List of Major League Baseball career assists leaders - Wikipedia
URL: https://en.wikipedia.org/.../List_of_Major_League_Baseball_career_assists_l...
Content Summary:
 The second baseman often possesses quick <b>hands</b> and feet, needs the ability to ... <b>Pat Burrell</b>, 92, 0, 92. Hank Sauer, 92, 0, 92. 46, Joe Vosmik, 90, 89, 1. 47&nbsp;...
... (content truncated to save space)
\end{lstlisting}
\vspace{0.5em}\hrule\vspace{0.5em}

\textbf{Assistant:}

Tool Calls:
query\_on\_page(url=https://en.wikipedia.org/wiki/Pat\_Burrell, question=What is Pat Burrell's batting hand?)
\vspace{1em}

\textbf{Tool Response (query\_on\_page):}

\begin{lstlisting}
Pat Burrell batted right-handed.
\end{lstlisting}
\vspace{0.5em}\hrule\vspace{0.5em}

\textbf{Assistant:}

Tool Calls:
search(query=Mark Mulder batting hand)
\vspace{1em}

\textbf{Tool Response (search):}

\begin{lstlisting}
1. Title: Mark Mulder - Wikipedia
URL: https://en.wikipedia.org/wiki/Mark_Mulder
Content Summary:
 <b>Mark</b> Alan <b>Mulder</b> (born August 5, 1977) is an American former professional <b>baseball</b> player. A left-handed starting pitcher, <b>Mulder</b> pitched in Major League&nbsp;...

2. Title: 2006 St. Louis Cardinals season - Wikipedia
URL: https://en.wikipedia.org/wiki/2006_St._Louis_Cardinals_season
Content Summary:
 Momentum would be slowed by injuries, as starting pitcher <b>Mark Mulder</b> was ... <b>Batting</b>. edit. Note: G = Games played; AB = At <b>bats</b>; R = Runs; H = Hits; 2B&nbsp;...

3. Title: Athletics (baseball) - Wikipedia
URL: https://en.wikipedia.org/wiki/Athletics_(baseball)
Content Summary:
 ... mark when a player <b>bats</b>. Prior to the mid-2010s, the A had a long ... <b>Mark Mulder</b>; (content truncated to save space)
\end{lstlisting}
\vspace{0.5em}\hrule\vspace{0.5em}

\textbf{Assistant:}

Tool Calls:
query\_on\_page(url=https://en.wikipedia.org/wiki/Mark\_Mulder, question=What is Mark Mulder's batting hand?)
\vspace{1em}

\textbf{Tool Response (query\_on\_page):}

\begin{lstlisting}
Mark Mulder's batting hand is left.
\end{lstlisting}
\vspace{0.5em}\hrule\vspace{0.5em}

\textbf{Assistant:}

Tool Calls:
search(query=Corey Patterson batting hand)
\vspace{1em}

\textbf{Tool Response (search):}

\begin{lstlisting}
1. Title: Corey Patterson - Wikipedia
URL: https://en.wikipedia.org/wiki/Corey_Patterson
Content Summary:
 Atlanta, Georgia, U.S.. <b>Batted</b>: Left. Threw: Right. MLB debut. September 18, 2000, for the Chicago Cubs.

2. Title: All-Star Final Vote - Wikipedia
URL: https://en.wikipedia.org/wiki/All-Star_Final_Vote
Content Summary:
 On Monday July 7, 2003, Kenny Lofton was added to the ballot to replace the Chicago Cubs&#39; <b>Corey Patterson</b> who was injured on the day before. ... <b>hand</b> and wearing&nbsp;...

3. Title: Cody Ross - Wikipedia
URL: https://en.wikipedia.org/wiki/Cody_Ross
Content Summary:
 He is one of the few Major League players to <b>bat</b> right-handed and throw left-handed. Cody Ross. Ross with the San Francisco Giants in 2011. Outfielder.
... (content truncated to save space)
\end{lstlisting}
\vspace{0.5em}\hrule\vspace{0.5em}

\textbf{Assistant:}
Tool Calls:
query\_on\_page(url=https://en.wikipedia.org/wiki/Corey\_Patterson, question=What is Corey Patterson's batting hand?)
\vspace{1em}

\textbf{Tool Response (query\_on\_page):}

\begin{lstlisting}
Corey Patterson's batting hand is left.
\end{lstlisting}
\vspace{0.5em}\hrule\vspace{0.5em}

\textbf{Assistant:}

Tool Calls:
search(query=Jeff Austin batting hand)
\vspace{1em}

\textbf{Tool Response (search):}

\begin{lstlisting}
1. Title: Austin Hedges - Wikipedia
URL: https://en.wikipedia.org/wiki/Austin_Hedges
Content Summary:
 <b>Austin</b> Charles Hedges (born August 18, 1992) is an American professional <b>baseball</b> catcher for the Cleveland Guardians of Major League <b>Baseball</b> (MLB).

2. Title: List of Major League Baseball batting champions - Wikipedia
URL: https://en.wikipedia.org/.../List_of_Major_League_Baseball_batting_champ...
Content Summary:
 In <b>baseball</b>, <b>batting</b> average (AVG) is a measure of a batter&#39;s success rate in achieving a hit during an at <b>bat</b>. In Major League <b>Baseball</b> (MLB),&nbsp;...

3. Title: Austin Riley - Wikipedia
URL: https://en.wikipedia.org/wiki/Austin_Riley
Content Summary:
 Memphis, Tennessee, U.S.. <b>Bats</b>: Right. Throws: Right. MLB debut. May 15, 2019, for the Atlanta Braves.
... (content truncated to save space)
\end{lstlisting}
\vspace{0.5em}\hrule\vspace{0.5em}

\textbf{Assistant:}

Tool Calls:
query\_on\_page(url=https://en.wikipedia.org/wiki/Jeff\_Austin, question=What is Jeff Austin's batting hand?)
\vspace{1em}

\textbf{Tool Response (query\_on\_page):}

\begin{lstlisting}
The content provided does not mention Jeff Austin's batting hand. However, it does identify Jeff Austin as a "American Major League Baseball pitcher," which implies he was a baseball player. Pitchers typically do not bat, so the question about his batting hand might not be applicable. If you are referring to a specific Jeff Austin in the context of baseball, additional information would be needed to determine his batting hand.
\end{lstlisting}
\vspace{0.5em}\hrule\vspace{0.5em}

\textbf{Assistant:}

Tool Calls:
search(query=J.D. Drew batting hand)
\vspace{1em}

\textbf{Tool Response (search):}

\begin{lstlisting}
1. Title: J. D. Drew - Wikipedia
URL: https://en.wikipedia.org/wiki/J._D._Drew
Content Summary:
 On August 9, on what would have been <b>Drew&#39;s</b> first game in Philadelphia, he sat out, citing a bruised right <b>hand</b>. ... <b>Drew</b> had a very good season, <b>batting</b> .284&nbsp;...

2. Title: Home run - Wikipedia
URL: https://en.wikipedia.org/wiki/Home_run
Content Summary:
 <b>J. D. Drew</b> has been part of two different sets of back-to-back-to-back-to ... &#39;Ralph,&#39; he said, &#39;the right-<b>hand batters</b> who hit to the right field are driving&nbsp;...

3. Title: Adam Wainwright - Wikipedia
URL: https://en.wikipedia.org/wiki/Adam_Wainwright
Content Summary:
 After the 2003 season, the Braves traded him to the Cardinals for outfielder <b>J. D. Drew</b>. ... <b>batting</b> practice on February 21. Three days later, the&nbsp;...



... (content truncated to save space)
\end{lstlisting}
\vspace{0.5em}\hrule\vspace{0.5em}

\textbf{Assistant:}

Tool Calls:
query\_on\_page(url=https://en.wikipedia.org/wiki/J.\_D.\_Drew, question=What is J. D. Drew's batting hand?)
\vspace{1em}

\textbf{Tool Response (query\_on\_page):}

\begin{lstlisting}
J. D. Drew's batting hand is left.
\end{lstlisting}
\vspace{0.5em}\hrule\vspace{0.5em}

\textbf{Assistant:}

Tool Calls:
search(query=Jeff Austin batting hand 1998 MLB draft)
\vspace{1em}

\textbf{Tool Response (search):}

\begin{lstlisting}
1. Title: 1998 Major League Baseball draft - Wikipedia
URL: https://en.wikipedia.org/wiki/1998_Major_League_Baseball_draft
Content Summary:
 The <b>1998</b> Major League <b>Baseball draft</b>, was the choosing of high school and college <b>baseball</b> players, held on June 2 and 3, <b>1998</b>. A total of 1445 players were&nbsp;...

2. Title: List of Boston Red Sox first-round draft picks - Wikipedia
URL: https://en.wikipedia.org/.../List_of_Boston_Red_Sox_first-round_draft_pic...
Content Summary:
 The Boston Red Sox are a <b>Major League Baseball</b> (<b>MLB</b>) franchise based in Boston, Massachusetts. They play in the American League East division.

3. Title: List of Detroit Tigers first-round draft picks - Wikipedia
URL: https://en.wikipedia.org/wiki/List_of_Detroit_Tigers_first-round_draft_picks
Content Summary:
 The Detroit Tigers are a <b>Major League Baseball</b> (<b>MLB</b>) franchise based in Detroit, Michigan. They play in the American League Central division.

... (content truncated to save space)
\end{lstlisting}
\vspace{0.5em}\hrule\vspace{0.5em}

\textbf{Assistant:}

Tool Calls:
query\_on\_page(url=https://en.wikipedia.org/wiki/Jeff\_Austin\_(baseball), question=What is Jeff Austin's batting hand?)
\vspace{1em}

\textbf{Tool Response (query\_on\_page):}

\begin{lstlisting}
Jeff Austin's batting hand is **Right**.
\end{lstlisting}
\vspace{0.5em}\hrule\vspace{0.5em}

\textbf{Assistant:}

The information retrieved confirms that the batting hands of the first five picks in the 1998 MLB draft are:

1. **Pat Burrell**: Right-handed
2. **Mark Mulder**: Left-handed
3. **Corey Patterson**: Left-handed
4. **Jeff Austin**: Right-handed
5. **J. D. Drew**: Left-handed

This matches the statement provided.

Final answer: Correct.

So the answer is The batting hands of the first five picks in the 1998 MLB draft are as follows:Pat Burrell: Right-handed, Mark Mulder: Left-handed, Corey Patterson: Left-handed, Jeff Austin: Right-handed, J. D. Drew: Left-handed.

\vspace{0.5em}\hrule\vspace{0.5em}

\textbf{Final Answer:}

The batting hands of the first five picks in the 1998 MLB draft are as follows:Pat Burrell: Right-handed, Mark Mulder: Left-handed, Corey Patterson: Left-handed, Jeff Austin: Right-handed, J. D. Drew: Left-handed.

\end{tcolorbox}

\begin{tcolorbox}[breakable, fontupper=\small,title=Case study 2.1 (WebSeer-14B)]
\textbf{System:}

You are a reasoning assistant with the ability to perform web searches and execute Python code to help you process the content of the page and answer the user's question accurately.
Do not use any knowledge you know; all facts in your thinking should be obtained from the information returned by the tools. You can repeat the search process multiple times if necessary. 
Once you have all the information you need, continue your reasoning.
Please first make a plan before calling tools.
Please answer the following question. You should provide your final answer to the "submit\_answer" tool.

\vspace{1em}

\textbf{User:}

Question: What German medium tank, sometimes called the "Mark V", that was deployed on the Eastern and Western fronts from mid-1943 provided the armor and suspension for the chassis of the tank destroyer built by Nazi Germany during World War II?
\vspace{1em}

\textbf{Assistant:}

Tool Calls:
search(query=German medium tank Mark V WWII)
\vspace{1em}

\textbf{Tool Response (search):}

\begin{lstlisting}
1. Title: Panther tank - Wikipedia
URL: https://en.wikipedia.org/wiki/Panther_tank
Content Summary:
 The Panther tank, officially Panzerkampfwagen <b>V</b> Panther (abbreviated Pz.Kpfw. <b>V</b>) with ordnance inventory designation: Sd.Kfz. 171, is a <b>German medium tank</b>&nbsp;...

2. Title: Mark V tank - Wikipedia
URL: https://en.wikipedia.org/wiki/Mark_V_tank
Content Summary:
 There are eleven surviving <b>Mark V tanks</b>. The Mark VIII <b>tank</b> was an enlarged <b>Mark V</b> with greater power: only those with the Liberty engine saw post-war service&nbsp;...

3. Title: German tanks in World War II - Wikipedia
URL: https://en.wikipedia.org/wiki/German_tanks_in_World_War_II
Content Summary:
 List of <b>tanks</b> &middot; Grosstraktor &middot; Leichttraktor &middot; Panzer I &middot; Panzer II &middot; Panzer III &middot; Panzer <b>IV</b> &middot; Neubaufahrzeug &middot; Panther.
... (content truncated to save space)
\end{lstlisting}
\vspace{0.5em}\hrule\vspace{0.5em}

\textbf{Assistant:}

Tool Calls:
query\_on\_page(url=https://en.wikipedia.org/wiki/Panther\_tank, question=What German medium tank, sometimes called the 'Mark V', was deployed on the Eastern and Western fronts from mid-1943?)
\vspace{1em}

\textbf{Tool Response (query\_on\_page):}

\begin{lstlisting}
The German medium tank, sometimes called the 'Mark V', that was deployed on the Eastern and Western fronts from mid-1943 is the **Panther tank**, officially designated as the **Panzerkampfwagen V Panther**.
\end{lstlisting}
\vspace{0.5em}\hrule\vspace{0.5em}

\textbf{Assistant:}

Tool Calls:
query\_on\_page(url=https://en.wikipedia.org/wiki/Panther\_tank, question=Which tank destroyer built by Nazi Germany during World War II used the Panther tank's armor and suspension for its chassis?)
\vspace{1em}

\textbf{Tool Response (query\_on\_page):}

\begin{lstlisting}
The tank destroyer built by Nazi Germany during World War II that used the Panther tank's armor and suspension for its chassis was the **Jagdpanther**. 

The Jagdpanther was a heavy Jagdpanzer (tank destroyer) that was based on the Panther tank's chassis, incorporating its armor and suspension system. It was equipped with an 88 mm long-barrel gun, and was designed to provide superior armor protection and mobility compared to other German tank destroyers of the time.
\end{lstlisting}
\vspace{0.5em}\hrule\vspace{0.5em}

\textbf{Assistant:}

The German medium tank, sometimes called the "Mark V," that was deployed on the Eastern and Western fronts from mid-1943 is the Panther tank (officially Panzerkampfwagen V Panther). The tank destroyer built by Nazi Germany during World War II that used the Panther tank's armor and suspension for its chassis is the Jagdpanther.

Final answer: Panther tank.

So the answer is Panther tankWait, I need to verify the answer:

Tool Calls:
search(query=German medium tank Mark V WWII)
\vspace{1em}

\textbf{Tool Response (search):}

\begin{lstlisting}
1. Title: Panther tank - Wikipedia
URL: https://en.wikipedia.org/wiki/Panther_tank
Content Summary:
 The Panther tank, officially Panzerkampfwagen <b>V</b> Panther (abbreviated Pz.Kpfw. <b>V</b>) with ordnance inventory designation: Sd.Kfz. 171, is a <b>German medium tank</b>&nbsp;...

2. Title: Mark V tank - Wikipedia
URL: https://en.wikipedia.org/wiki/Mark_V_tank
Content Summary:
 There are eleven surviving <b>Mark V tanks</b>. The Mark VIII <b>tank</b> was an enlarged <b>Mark V</b> with greater power: only those with the Liberty engine saw post-war service&nbsp;...

3. Title: German tanks in World War II - Wikipedia
URL: https://en.wikipedia.org/wiki/German_tanks_in_World_War_II
Content Summary:
 List of <b>tanks</b> &middot; Grosstraktor &middot; Leichttraktor &middot; Panzer I &middot; Panzer II &middot; Panzer III &middot; Panzer <b>IV</b> &middot; Neubaufahrzeug &middot; Panther.
... (content truncated to save space)
\end{lstlisting}
\vspace{0.5em}\hrule\vspace{0.5em}

\textbf{Assistant:}

Tool Calls:
query\_on\_page(url=https://en.wikipedia.org/wiki/Panther\_tank, question=Was the Panther tank deployed on the Eastern and Western fronts from mid-1943?)
\vspace{1em}

\textbf{Tool Response (query\_on\_page):}

\begin{lstlisting}
Yes, the Panther tank was deployed on both the Eastern and Western fronts from mid-1943. On the Eastern Front, it was introduced during the Battle of Kursk in the summer of 1943, and it was used throughout the war. On the Western Front, it was deployed in France, with the first Panther-equipped Panzer regiments arriving there in June 1944. The tank served in most European theatres of World War II from mid-1943 to the end of the war in May 1945.
\end{lstlisting}
\vspace{0.5em}\hrule\vspace{0.5em}

\textbf{Assistant:}

Tool Calls:
query\_on\_page(url=https://en.wikipedia.org/wiki/Jagdpanther, question=What tank destroyer used the Panther tank's armor and suspension for its chassis?)
\vspace{1em}

\textbf{Tool Response (query\_on\_page):}

\begin{lstlisting}
The tank destroyer that used the Panther tank's armor and suspension for its chassis is the **Jagdpanther**.
\end{lstlisting}
\vspace{0.5em}\hrule\vspace{0.5em}

\textbf{Assistant:}

The Panther tank was indeed deployed on both the Eastern and Western fronts from mid-1943, and the tank destroyer that used its armor and suspension for its chassis is the Jagdpanther.

Final answer: Correct So the answer is Panther tank

Tool Calls:
submit\_answer(answer=Panther tank)

\vspace{0.5em}\hrule\vspace{0.5em}

\textbf{Final Answer:}

Panther tank

\end{tcolorbox}
\section{The usage of LLM}
Portions of this paper were polished by the large language models (LLMs), which were used to improve the clarity, grammar, and presentation of the text. The models were not used to generate research ideas, conduct experiments, or analyze results; all conceptual contributions and empirical findings are the work of the authors. We carefully reviewed and edited all LLM-generated suggestions to ensure accuracy and alignment with the intended meaning.

\end{document}